\documentclass{ecai} 

\usepackage{latexsym}
\usepackage{amssymb}
\usepackage{amsmath}
\usepackage{amsthm}
\usepackage{booktabs}
\usepackage{enumitem}
\usepackage{graphicx}
\usepackage{color}
\usepackage{listings}
\usepackage{times}
\usepackage{soul}
\usepackage{url}
\usepackage[hidelinks]{hyperref}
\usepackage[utf8]{inputenc}
\usepackage[small]{caption}
\usepackage{graphicx}
\usepackage{amsmath}
\usepackage{amsthm}
\usepackage{booktabs}
\usepackage[switch]{lineno}
\usepackage{graphicx}  
\usepackage{svg}  
\usepackage{multirow}
\usepackage{float}
\usepackage{CJK}
\usepackage{algorithm,algpseudocode}
\usepackage{graphicx}
\usepackage{amsmath}
\usepackage{soul}
\usepackage{xcolor}

\usepackage{color}
\usepackage{times}
\usepackage{soul}
\usepackage{url}
\usepackage[hidelinks]{hyperref}
\usepackage[utf8]{inputenc}
\usepackage[font=small]{caption}
\usepackage{amsmath}
\usepackage{amsthm}
\usepackage{booktabs}
\usepackage{algpseudocode}
\usepackage[switch]{lineno}
\usepackage{multirow}
\usepackage{float}
\usepackage{CJK}
\usepackage{graphicx}
\usepackage{amsmath}
\usepackage{soul}
\usepackage{xcolor}
\usepackage{cleveref}
\usepackage{appendix}

\newcommand{\BibTeX}{B\kern-.05em{\sc i\kern-.025em b}\kern-.08em\TeX}

\begin{document}


\begin{frontmatter}

\title{DatasetAgent: A Novel Multi-Agent System \\for Auto-Constructing Datasets from Real-World Images}
\author[A,B]{\fnms{Haoran}~\snm{Sun}\thanks{Email: 2022090916002@std.uestc.edu.cn.}\footnote{Equal contribution.}}
\author[A,B]{\fnms{Haoyu}~\snm{Bian}\thanks{Email: 2022090914006@std.uestc.edu.cn.}\footnotemark[1]}
\author[A,C]{\fnms{Shaoning}~\snm{Zeng}\thanks{Corresponding Author. Email: zeng@csj.uestc.edu.cn.}}
\author[B]{\fnms{Yunbo}~\snm{Rao}\thanks{Corresponding Author. Email: raoyb@uestc.edu.cn.}}
\author[D]{\fnms{Xu}~\snm{Xu}\thanks{Email: 051020002@fudan.edu.cn.}}
\author[D]{\fnms{Lin}~\snm{Mei}\thanks{Email: mei\_lin@vip.126.com.}}
\author[E]{\fnms{Jianping}~\snm{Gou}\thanks{Email: pjgzy61@swu.edu.cn.}}

\address[A]{Yangtze Delta Region Institute (Huzhou), University of Electronic and Science Technology of China, Huzhou, Zhejiang, China}
\address[B]{School of Information and Software Engineering, University of Electronic and Science Technology of China, Chengdu, Sichuan, China}
\address[C]{Zhejiang Chuangjiekedong Ltd., Huzhou, Zhejiang, China}
\address[D]{DongHai Laboratory, Zhoushan, Zhejiang, China}
\address[E]{College of Computer and Information Science, Southwest University, China}

\begin{abstract}
Common knowledge indicates that the process of constructing image datasets usually depends on the time-intensive and inefficient method of manual collection and annotation. Large models offer a solution via data generation. Nonetheless, real-world data are obviously more valuable comparing to artificially intelligence generated data, particularly in constructing image datasets. For this reason, we propose a novel method for auto-constructing datasets from real-world images by a multiagent collaborative system, named as DatasetAgent. By coordinating four different agents equipped with Multi-modal Large Language Models (MLLMs), as well as a tool package for image optimization, DatasetAgent is able to construct high-quality image datasets according to user-specified requirements. In particular, two types of experiments are conducted, including expanding existing datasets and creating new ones from scratch, on a variety of open-source datasets. In both cases, multiple image datasets constructed by DatasetAgent are used to train various vision models for image classification, object detection, and image segmentation. 

\end{abstract}
\end{frontmatter}

\section{Introduction}

Thanks to the rapid development of Large Language Models (LLMs) \cite{cao2023comprehensive,roser2024brief,li2024personal} and Multi-Modal Large Language Model \cite{wu2023brief,digiorgio2023artificial}, Artificial Intelligence (AI) Agent \cite{zeng2023agenttuning} becomes one of the most prevailing technologies. In particular, LLMs are equipped with extensive training parameters and complex deep neural networks, and capable of emulating human responses to natural language queries and handling highly intricate tasks and data. While MLLMs integrate the capabilities of understanding and processing multi-model data including text, voice, images and videos \cite{wu2023brief}. AI Agents consist of software components that execute actions on behalf of a user or another program, so that it can observe and respond to its environment \cite{zeng2023agenttuning}. It has a wide applications across many different industries. Based on these foundational models, AI Agents facilitate the interaction among multiple modals of data, and therefore autonomously perform specialized or complex work tasks. In this way, AI Agents help reduce human labor in certain tasks, lighten the human workload, and enhance efficiency across various domains, especially in providing a rapid dataset foundation for AI-enabled technologies applications.

\begin{figure}[H]
  \centering
  \includegraphics[width=\columnwidth]{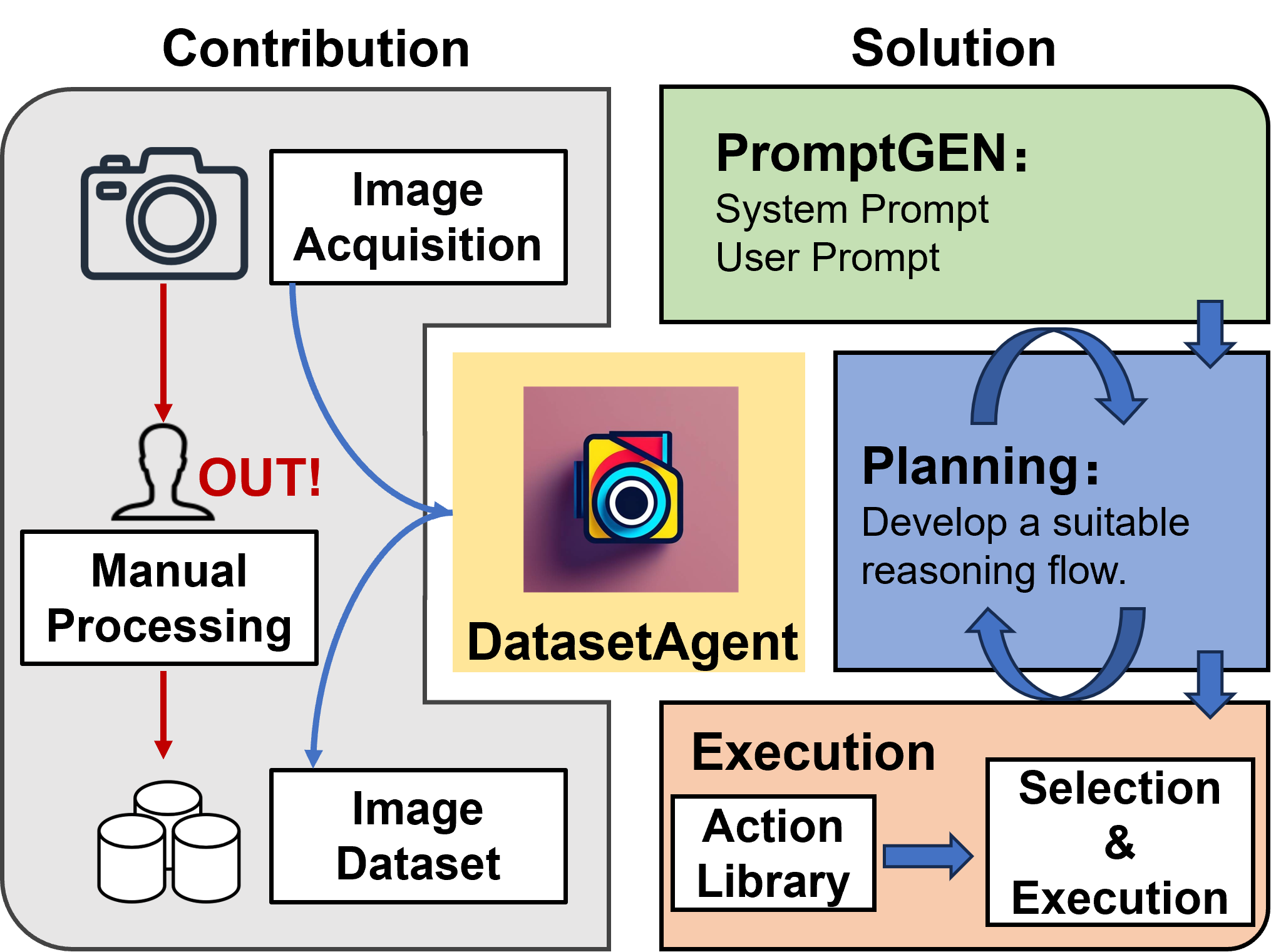}
  \caption{DatasetAgent replaces traditional manual methods for constructing real-world image datasets. By analyzing multiple inputs, it autonomously plans and executes the process for constructing custom datasets. 
  }
  \label{fig:first}
\end{figure}

The application of AI Agents is increasingly expanding over time, particularly in areas such as natural language dataset construction \cite{abdullin2024synthetic,isazawa2023automated,wang2021data}. However, to the best of our knowledge, their utilization in the construction of image datasets is still unexplored. The process of training deep learning networks using image datasets has a great impact on the result of image classification \cite{bansal2023transfer}. Therefore, numerous researchers have constructed many image datasets for this purpose, e.g., ImageNet \cite{deng2009imagenet}, CIFAR-10 
\cite{krizhevsky2009learning}, STL-10 \cite{coates2011analysis}, etc. Additionally, to address complex real-world applications like autonomous driving \cite{yurtsever2020survey} and industrial inspection \cite{liu2023survey}, a series of object detection and image segmentation datasets, such as MS-COCO \cite{lin2014microsoft}, PASCAL VOC \cite{everingham2015pascal}, Cityscapes \cite{cordts2016cityscapes}, and CamVid \cite{brostow2008segmentation}, have also been continuously proposed and developed, which emphasizes the importance of image datasets.

Traditionally, constructing image datasets involves either manual collection, processing, and annotation \cite{7882707,kawano2015automatic,alkhalid2022expansion}, or the automatic image generation or synthesis \cite{barrera2023generating,soltani2016automated}. Indeed, the process of manual construction is both laborious and ineffective, and therefore failing to attain in a way of complete automation. The deployment of generated or synthetic images inadequately accounts for diverse viewpoints, illumination, and world conditions, resulting in a dataset quality that is inferior to one constructed with real-world photographs. 
 However, most of the contemporary AI Agents still concentrated on language-oriented reasoning, offering no comprehensive solution for auto-constructing an image dataset. 


To address the gap in the application of AI Agents based on LLMs for image dataset construction, 
we develop a Multi-Modal multiagent collaborative system, named as DatasetAgent, capable of automatically constructing image datasets. Compared to traditional methods heavily depending on human labor, DatasetAgent requires only real-world images or image resources, as well as a brief description of target dataset requirement. Then, it automatically handles all the subsequent operations, including image collection (optional), image processing, cleaning, and annotation. This multiagent system is developed through a strategic combination of prompts and MLLMs, working in harmony to enhance their collective function. The main purpose of DatasetAgent is to improve the efficiency of constructing image datasets and reduce human labor. While effectively liberating human labor, the quality of the constructed image dataset can also be guaranteed. Moreover, it employs real-world images to overcome the inherent deficiencies associated with the generated or synthetic images.

The contributions of this work are as follows. The completely automated construction of high-quality image datasets in a modern AI Agent, DatasetAgent, is presented. Demonstrating its effectiveness in two cases of image dataset construction, i.e., augmenting existing datasets and creating new ones from scratch using a variety of online open-source datasets. Experimental results demonstrate a consistent improvement of downstream models on these constructed datasets, and an average accuracy up to 98.90\% is obtained. We also validated the high quality of six image classification datasets constructed by DatasetAgent using six evaluation metrics.


\section{Related Work}
\subsection{AI Agent}

AI Agent is defined as an autonomous entity capable of executing assigned tasks \cite{zeng2023agenttuning}. This concept was introduced by AI researchers to emulate human intelligence. These AI-powered agents can self-generate and execute tasks, continuously innovate, reprioritize task queues, and engage in high-level tasks, all in a cyclical effort to achieve their goals. Modern frameworks like AgentLite \cite{liu2024agentlite} integrate Large Language Models (LLMs), planning, memory, and tools, allowing for task automation within the LLM context and fostering ongoing exploration, strateging, and skill development \cite{wang2024survey,xi2023rise}. 

With the advancement of artificial intelligence technologies, agents combining LLMs with deep reinforcement learning have become a key component in AI applications \cite{wang2024survey,xi2023rise}. They are making a great impact in fields like intelligent systems \cite{zhang2024agentcf}, robotics \cite{bousmalis2023robocat}, gaming \cite{gong2023mindagent}, automation \cite{abras2008multi}, and data analysis \cite{bose1999application}. Agents based on LLMs have successfully generated training data and automated dataset construction for training language models. For instance, a method for generating sample dialogues for developing and training conversational agents was proposed, utilizing prompt engineering to develop two agents that ``converse'' with each other. This method leveraged OpenAI's ChatGPT4 and an agent to create a dialogue dataset for training \cite{abdullin2024synthetic}. $Art\_GenEvalGPT$, a new dataset of synthetic dialogues focused on art, was generated through ChatGPT. It explored nuanced conversations about artworks, artists, and genres, incorporating emotional interventions and subjective opinions in various roles, like teacher-student or expert guides \cite{gil2024dataset}. ParaGPT was a new paraphrasing dataset consisting of 81,000 machine-generated sentence pairs, among which 27,000 sentences were generated using ChatGPT. Based on evaluation metrics, the performance of ChatGPT was considered impressive \cite{kurt2024comparative}. 

As these technologies continue to advance, the integration of AI Agents with MLLMs has become increasingly tighter. The capabilities of AI Agents in handling multi-modal tasks, such as in image processing, are growing stronger. However, to our understanding, in image dataset construction, MLLMs and Agents have not been successfully applied. How to have a sophisticated design of Agent for auto-constructing an image dataset remains unexplored.

\begin{figure*}[h]
    \centering
    \includegraphics[width=0.9\linewidth]{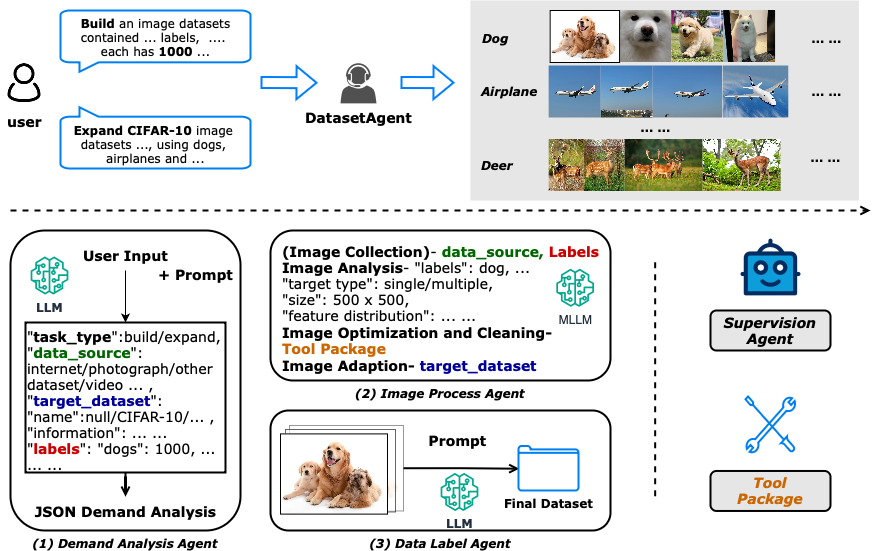}
    \caption{Overview of the DatasetAgent workflow for auto-constructing image classification datasets. The process begins with the user's requirements, then multiple agents work together to complete the fully automated dataset construction. Users can choose to build a new image dataset or expand an existed dataset, using real-world image data provided by users or automatically collected by DatasetAgent. Ultimately, a successfully constructed new image dataset is output.}
    \label{fig:second}
\end{figure*}


\subsection{Dataset Construction}

Traditionally, constructing an image dataset heavily depended on manual labor, requiring a great mass of human involvement in collecting image data, adjusting image size and resolution, image segmentation with annotating bounding boxes for object detection, performing pixel-level labeling for segmentation tasks, and other processing \cite{song2019survey,minaee2021image,jiao2019survey}. The manual merging and annotation validation of images into the dataset must also follow strict standards \cite{deng2009imagenet}. Semi-automatic methods can utilize machine assistance for data retrieval,  image importation or assisted annotation, but still require significant human involvement \cite{jiang2023review,7882707,kawano2015automatic,alkhalid2022expansion}.

In addition to these two aforementioned methods, another approach is direct generation or synthesis of image data. For example, an automated method for creating and annotating synthetic datasets of construction equipment images was developed \cite{soltani2016automated}. They combined 3D models of machinery with background images from construction sites. 


The other two methods were proposed for generating synthetic construction image datasets. The first one focused on single construction assets in the camera view, while the second included multiple assets, allowing for occlusion or partial views from different angles \cite{barrera2023generating}. 

However, overwhelming manual operations in constructing a dataset makes it hard to have an efficient automation tool. When constructing an image dataset, one must follow a set of strict requirements and specifications on objectives, size, dimensions, and resolution. The images retrieved by machines cannot perfectly meet these standards. Human operations were still needed to rigorously select, adjust, and annotate the collected images. Unlike image classification tasks, constructing object detection datasets requires dual annotation of the location and category of each object in the image.  This is typically done via specific annotation formats containing bounding box or keypoint coordinates, such as YOLO \cite{redmon2016you}, VOC \cite{everingham2010pascal} and COCO \cite{lin2014microsoft} formats.Building image segmentation datasets is more demanding, requiring pixel - level annotation.  Semantic segmentation needs category labels assigned to each pixel, while instance segmentation further requires distinguishing different instances of the same category.  This annotation relies on professional tools like LabelMe \cite{russell2008labelme} and COCO Annotator \cite{stefanics2022coco} for mask generation, with time and cost possibly reaching dozens of times that of image classification annotation. Therefore, even with machine assistance in constructing image datasets, the entire process remains limited to manual or semi-automated procedures. The result is requiring substantial human involvement in the annotation process, and failing to achieve full automation. What is worse, a significant consumption of human resources was required and the process was low efficiency. 

On the other hand, if images are generated or synthesized directly, many influencing factors such as the perspective, lighting, and weather conditions of the images cannot be well accommodated \cite{barrera2023generating}. Consequently, the quality of the datasets created in this generation way is far behind the ones created from real-world images. This is why most model training still relies on data collection from commercial data providers like Scale AI. Besides this, many LLM providers like OpenAI pays attention to develop their own technique for detection of generated content, to avoid collecting generating data \cite{li2024mage}.In the experiments on image datasets, we observed that the performance of the downstream models is impacted.

Currently, there is no methodology capable of completely automating the construction of image datasets, nor the processing and seamless integration of captured or recorded real-world images into these datasets. This is the very motivation that leads to have our proposed DatasetAgent. 

\section{DatasetAgent}

\subsection{Architecture}

DatasetAgent builds multiple agents with different responsibilities to work together, namely Demand Analysis Agent, Image Processing Agent, and Data Label Agent respectively. They are monitored by Supervision Agents to achieve fully automatic construction of image datasets for different tasks. Firstly, the Demand Analysis Agent conducts the requirement analysis based on user input, and then the Image Process Agent executes image collection (optional), image analysis, image optimization and image clearing, and adjusts them to meet the requirements of the target dataset. At the same time, we also provide a Tool Package that can be called by the Image Process Agent, which includes various image processing tool interfaces. The Image Process Agent decides whether to use and how to use it to optimize the quality of the target dataset by itself. Finally, the Data Label Agent analyzes, annotates, filters, and matches the information obtained by the first two Agents, and puts the processed image data into the corresponding label of the target dataset. During the process of dataset construction, the Supervision Agent monitors and logs the status of these three agents throughout the whole process. When problems such as generating illusions or tool errors occur, Supervision Agent will solve the problem after reasoning, analysis, and reflection to ensure that DatasetAgent can smoothly construct a high-quality image dataset. Table \ref{tab:case} shows an example of workflow during the construction.

\begin{table*}[h]
  \caption{An example workflow of DatasetAgent.}
  \label{tab:case}
  \centering
  \begin{tabular}{cclll}
    \toprule
    \multirow{1}*{\textbf{User}}  & \multirow{1}*{\textbf{Input}} 
    & \multicolumn{1}{l}{I want to expand an image dataset called CIFAR-10. Add 5000 images each label.} 
    \\
    
    \midrule
    \multirow{2}*{\textbf{Demand Analysis}}  & \textbf{Demand Analysis Agent} 
    & \multicolumn{3}{l}{Analyze user input and extract expansion requirement: [CIFAR-10, +5000 images/class]}\\ 
    & return & \multicolumn{3}{l}{LLM: $\{$"User Demand":...., "Target Dataset": ...$\}$} \\ 
    
    \midrule   
    \multirow{2}*{\textbf{Image Collection}}  & \textbf{Image Process Agent} 
    & \multicolumn{3}{l}{Begin image collection based on keywords and user-defined concepts.} \\ 
    & return & \multicolumn{3}{l}{Image Collected: \textbf{50498} raw images.} \\
    
    \midrule
    \multirow{3}*{\textbf{Image Analysis}}  & \textbf{Image Process Agent} 
    & \multicolumn{3}{l}{Analyzing collected images to extract semantic features and quality indicators.} \\
    & return & \multicolumn{3}{l}{Images 0001–2998 analyzed successfully.} \\
    & \textbf{Supervision Agent} & \multicolumn{3}{l}{\textcolor{red}{\textbf{[Error] Image \#2999: format decode failed (corrupt JPEG)}}} \\
    
    \midrule
    \multirow{3}*{\textbf{Error Recovery}} & \textbf{Supervision Agent} 
    & \multicolumn{3}{l}{Diagnose issue: corrupted image at \#2999. Log checkpoint saved.} \\
    & Action & \multicolumn{3}{l}{LLM: Skip corrupted image; alert Image Process Agent; restore from checkpoint \#2998.} \\
    & Resume & \multicolumn{3}{l}{Continue analysis from image \#3000.} \\
    
    \midrule
    \multirow{2}*{\textbf{Image Optimization \& Cleaning}}  & \textbf{Image Process Agent} 
    & \multicolumn{3}{l}{Crop object regions, remove redundant background, normalize quality.} \\
    & return & \multicolumn{3}{l}{LLM: All valid images processed and quality filtered (N=48912).} \\
    
    \midrule
    \multirow{2}*{\textbf{Image Adaption}}  & \textbf{Image Process Agent} 
    & \multicolumn{3}{l}{Align processed images with CIFAR-10 class definitions.} \\
    & return & \multicolumn{3}{l}{Images classified and grouped per label. Category balancing applied.} \\
    
    \midrule
    \multirow{2}*{\textbf{Data Annotation}}  & \textbf{Data Label Agent}
    & \multicolumn{3}{l}{Resize images to \textbf{32$\times$32}, annotate with class labels and instance IDs.}\\
    & return & \multicolumn{3}{l}{All images adapted to CIFAR-10 format. JSON metadata generated.} \\
    
    \midrule
    \multirow{2}*{\textbf{Final Verification}} & \textbf{Supervision Agent} 
    & \multicolumn{3}{l}{Cross-check label distributions, missing data, and image quality metrics.} \\
    & return & \multicolumn{3}{l}{\textbf{Passed.} Total 48912 new images added to CIFAR-10.} \\
    
    \midrule
    \multirow{1}*{\textbf{End}} & \textbf{Output} 
    & \multicolumn{3}{l}{DatasetAgent: successfully expanded CIFAR-10 with \underline{\textbf{48912}} high-quality images.}\\
    
    \bottomrule
\end{tabular}
\end{table*}

\subsection{Demand Analysis Agent}

Demand Analysis Agent interacts with large language models (LLMs) using carefully crafted prompts, equipping the LLM with the role of information extraction and semantic understanding to perform in-depth and structured analysis of user-provided requirements. Its primary objective is to extract essential components necessary for constructing image datasets, including the task type (classification, detection, segmentation), image source, and dataset specification. The system supports two task types: building a dataset from scratch (Build) with multi-task annotation configurations or expanding an existing one (Expand) while preserving original annotation types. Image sources may come from user-provided images, publicly available datasets, or, in the absence of an explicit source, the system will invoke the Image Process Agent to retrieve relevant images from the internet based on interpreted user intent. For dataset specifications, users constructing new datasets are expected to define structural and content-related expectations, while for dataset expansion, the system automatically retrieves metadata and design standards of the existing dataset to ensure consistency and usability. To address the temporal limitations of pre-trained LLMs, users may optionally supply related dataset papers or abstracts, thereby activating a retrieval-augmented generation (RAG) mechanism to enhance the accuracy and contextual relevance of the analysis. If the system detects that the user’s input is unrelated to the task of constructing a multi-task image dataset or lacks essential information, the Demand Analysis Agent will proactively issue prompts to the user, ensuring both the validity of the construction process and the completeness of the acquired information.

\subsection{Image Process Agent}

\textbf{Image Collection. }When the user does not specify the source of image data, the Image Process Agent autonomously crawls images from the internet based on the dataset objectives extracted by the Demand Analysis Agent. During the crawling process, the system first invokes a multimodal large language model (MLLM), guided by designed prompts, to conduct fine-grained analysis of candidate images. This analysis extracts rich visual and contextual information, including object category, appearance, background context, lighting conditions, camera angles, resolution, and fine-grained attributes. These attributes are then matched against the user requirements to identify and select suitable images for dataset construction, implementing cross-modal semantic alignment between images and user requests.. To ensure diversity and generalizability in the resulting dataset, the Image Process Agent continuously monitors the distribution of collected images and dynamically adjusts its crawling strategy. This includes tuning the number and types of images gathered, with a focus on enhancing intra-class diversity across visual, semantic, structural, and annotation dimensions. Such adaptive sampling ensures that images within each class are not only visually diverse but also semantically rich, which significantly contributes to the overall dataset quality. This approach improves the robustness and generalization performance of image classification models trained on the dataset and plays a key role in mitigating overfitting, a well-known challenge in deep learning, particularly in scenarios involving limited or homogeneous training data \cite{srivastava2014dropout}.\\
 \\
 
\noindent \textbf{Image Analysis, }powered by MLLM, performs fine-grained, instance-level analysis of each candidate image, extracting structured information from both visual and semantic perspectives, with all outputs strictly formatted in JSON, as shown in List \ref{imageanalysis}. Guided by natural language prompts, the MLLM is capable of generating detailed semantic representations of images, capturing key dimensions such as target category with instance counts, object appearance attributes (e.g., shape, color, texture) and spatial relationships, background composition, camera viewpoint, and lighting conditions. In addition, it evaluates image quality in terms of resolution, sharpness, color style, and completeness of local details. Crucially, the semantic analysis focuses on the extraction of fine-grained attributes including occlusion levels and boundary complexity, enabling the identification of subtle yet semantically significant variations within the same category—for instance, differences in morphology, material, or pose suitable for multi-task learning. These attributes play a pivotal role in enhancing the discriminative power of the dataset and improving the generalization ability of downstream models. The Image Process Agent compares the extracted semantic features of each image with the user-defined dataset requirements across multiple dimensions, employing a semantic similarity evaluation mechanism to assess their alignment with the specified acquisition criteria. To further support the robustness and usability of the resulting dataset, the system incorporates quality-aware analysis, identifying potential issues such as incomplete bounding boxes, occlusion affecting mask continuity, blurriness, exposure anomalies, or perspective distortion that may adversely affect learning outcomes, thereby facilitating subsequent data cleaning. \\
\\

\begin{lstlisting}[caption={Image Analysis Details}, label={imageanalysis}, escapeinside={(*@}{@*)}]
"image_id": "wild_fox_03821.jpg",
"target_category": "Fox",
"fine_grained_attributes": {
    "pose": "crouching,head turned to left",
    "pose_bounding_box":[0.22,0.35,0.75,0.65], 
    "fur_detail":"visible winter coat pattern",
    "fur_region": [0.25,0.4,0.7,0.6],
    "facial_features": {
    "snout_shape": "sharp and narrow",
    "eye_color": "dark amber",
    "ear_shape": "pointed with black tips",
    "facial_region": [0.6,0.38,0.72,0.48] },
    "tail_detail": {
        "appearance":"fluffy,white-tipped",
        "tail_region":[0.3,0.55,0.6,0.72]}}},
"background_composition": {
    "scene_type":"forest floor and leaf litter",
    "background_distribution": {
    "vegetation_area": [0.05,0.6,0.3,0.95],
    "ground_area": [0.0,0.75,1.0,1.0]},
"viewpoint_conditions": {
    "camera_angle": "frontal left,45",
    "camera_elevation": "eye-level",
    "lighting":"daylight with soft shadowing",
    "light_direction_vector": [-0.6,-0.4], 
    "depth":"strong focus on main subject"},
"image_quality": {
    "resolution": "1024x768",
    "sharpness_score": 0.94,
    "color_fidelity": "high",
    "detail_completeness": "98.7%",
    "style_consistency": "natural daylight",
    "jpeg_artifacts": false},
"semantic_alignment": {
    "class_prototype": "Fox",
    "similarity_score": 0.931,
    "match_features": [
    "fur color and texture",
    "tail shape and region",],
    "alignment_vector_diff": {
    "fur_texture": 0.02,
    "tail_geometry": 0.03,
    "facial_features": 0.01,
    "pose_alignment": 0.04},}},
"quality_risks": {"occlusion_detected": false,
    "blur_score": 0.06,
    "exposure_abnormality": false,
    "viewpoint_deviation_score": 0.08,
    "noise_level": "low",
    "warnings": [],
    "total_risk_score": 0.07},
"decision":{"qualified": true,
    "confidence": 0.982,
    "reason":"All required semantic features 
    and quality criteria met."}}

\end{lstlisting}

\noindent \textbf{Image Optimization and Cleaning. }Based on the integrated results of image analysis and user-defined requirements, the Image Process Agent first performs a comprehensive evaluation of the classification semantics, detection instances, and segmentation boundaries. In this process, the MLLM is responsible for inferring whether each image requires content-level optimization or attribute adjustment, and for determining specific processing strategies—including which images should undergo particular types of operations and which should be filtered out due to misalignment with task specifications. This decision-making is fully automated and executed by the MLLM, which, based on the analysis outcomes, selects appropriate tools from the predefined Tool Package and plans the invocation pathway. The system then automatically calls the corresponding tool interfaces according to the inference results, performs the necessary operations on the images, and completes the optimization workflow, thereby achieving fine-grained alignment between image properties and task requirements. Finally, combined with the information of the target dataset, all images are adjusted according to the size requirements of the target dataset, so that each image is consistent with the resolution requirements of the target dataset.

\subsection{Data Label Agent}

During the image optimization and processing performed by the Image Process Agent, the Data Label Agent operates in parallel to handle the final labeling and integration of processed images into the target dataset. This agent automatically matches each optimized image with its semantic information and classifies it into the appropriate label directory according to the predefined structure of the dataset. For tasks involving object detection and segmentation, the Data Label Agent identifies and annotates target objects by leveraging state-of-the-art Visual Language Models (VLMs) or Large Vision Models (LVMs). Users can select from models such as SAM 2 \cite{ravi2024sam2}, Grounding Dino 1.5 \cite{ren2024grounding}, and Grounded SAM \cite{ren2024grounded} to generate various annotation formats and corresponding mask images for semantic, instance, and panoptic segmentation tasks. The agent autonomously generates diverse prompts, including point prompts and box prompts in JSON format, as well as customized text prompts, while establishing high confidence thresholds (e.g., requiring a minimum confidence score of 0.5 for inclusion in the dataset) to ensure precise identification and segmentation. This comprehensive annotation process integrates both the target categories and fine-grained attributes extracted during image analysis with user-defined taxonomy and task-specific requirements, ultimately ensuring high accuracy, consistency, and semantic alignment in label assignment.

\begin{figure*}[h]
  \centering
  \includegraphics[width=\linewidth]{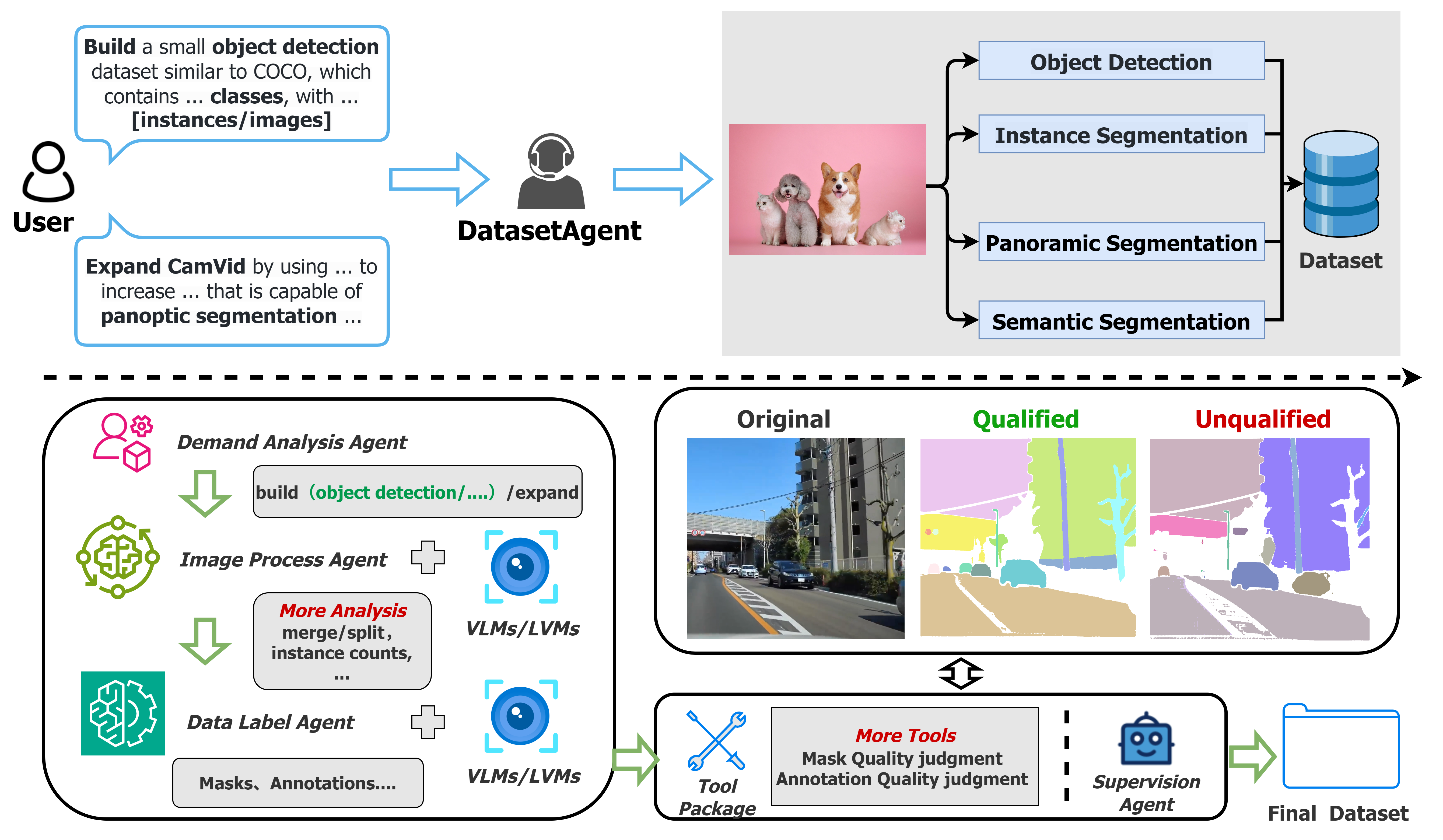} 
  \caption{Overview of the DatasetAgent workflow for auto-constructing image datasets(object detection and image segmentation). Based on the foundation in Figure \ref{fig:second}, the DatasetAgent can be designed to work with Visual Language Models (VLMs) or Large Vision Models (LVMs) for unsupervised annotation, enabling the construction of more diverse types of image datasets. The DatasetAgent drives VLMs/LVMs to annotate images and generate masks according to user requirements, while also evaluating the quality of annotation files.}
  \label{fig:VOC2007}
\end{figure*}

\subsection{Supervision Agent}

Throughout the collaborative workflow of the three primary agents, the Supervision Agent serves as the central module for monitoring, fault tolerance, and system coordination. It continuously tracks the execution status of each agent and records key process checkpoints, log data, and intermediate results into persistent storage to enable seamless recovery at any point. If any agent encounters an error—such as hallucinated outputs, tool invocation failures, or issues that obstruct the dataset construction pipeline—the Supervision Agent immediately captures and logs the error type and context. It then passes this information to a large language model (LLM) for diagnosis. Based on the analysis and repair strategy of LLM, the Supervision Agent autonomously performs error correction or agent-level restarts and restores the system to the last valid checkpoint, ensuring consistency and continuity after interruptions or crashes. Moreover, the Supervision Agent dynamically orchestrates agent execution and resource allocation based on real-time workload, operational status, and system efficiency. This enables parallelism, load balancing, and adaptive control across the entire data processing and labeling pipeline. Through its robust monitoring, automated recovery, and intelligent scheduling capabilities, the Supervision Agent ensures the DatasetAgent system can efficiently construct high-quality image datasets with reliability, resilience, and precision.

\subsection{Tool Package}

Tool Package is the core component responsible for supporting image optimization and cleaning operations. It encapsulates a collection of functional image processing modules that cover a wide range of common preprocessing tasks. Designed with modularity and composability in mind, this component enables the MLLM (Multimodal Large Language Model) to dynamically plan operation chains and accurately select appropriate tools based on image analysis results and user-defined requirements during inference. The Tool Package itself does not execute image processing tasks proactively; instead, it is invoked on demand under the guidance of MLLM decision-making, enabling targeted adjustment and optimization of image attributes. The MLLM outputs structured JSON data containing specific analysis metrics—such as the location of visual features in images requiring cropping—which are then parsed and mapped to corresponding tool interfaces. This facilitates the automatic execution of image processing pipelines in a fully autonomous manner.

The Tool Package supports a range of preprocessing operations. Size adjustment operations such as cropping and resizing help remove irrelevant regions and standardize image dimensions for normalized model input. Cropping can be automatically performed based on the detected location of target objects to improve the relevance of training data, while resizing employs interpolation methods—such as bilinear and bicubic interpolation—to ensure consistent visual quality across varying resolutions. For color adjustment, the Tool Package provides both color space conversion and color normalization functions to mitigate variations caused by lighting conditions and capture devices. Transforming images from the RGB color space to HSV or LAB enhances the expressiveness of color-related features, and normalization helps reduce color discrepancies across images to ensure dataset consistency. Normalization and standardization are also key preprocessing steps aimed at improving model training efficiency and robustness. By scaling pixel values to a standard range and standardizing data distribution (e.g., zero mean and unit variance), the model is better equipped to handle heterogeneous image inputs with numerical stability. In terms of data augmentation, the Tool Package supports commonly used techniques such as rotation, flipping, noise addition, and cropping with padding. These augmentations enhance dataset diversity and improve model generalization to different scenarios. For instance, rotation and flipping help the model learn features from objects in varying orientations, while noise injection increases robustness against real-world noise interference. Cropping and padding further enrich spatial diversity and maintain compatibility with model input size constraints. Meanwhile, for the generation of annotation files and masks, detection tools have also been set up to determine whether the generated annotation files conform to the format required by users, as well as whether the masked images have abnormal jagged edges and unreasonable vulnerabilities.

Through the integration of these operations, the Tool Package provides a robust and flexible framework for fine-grained image adjustment, ensuring that the processed images are optimally aligned with task requirements and capable of contributing to a high-quality, generalizable training dataset.

\section{Experiments}

\subsection{Setup}

This experiment is divided into two major stages: automatic construction of multi-task image datasets (classification, object detection, and segmentation including instance/semantic/panoptic types) and downstream validation. In the dataset construction stage, we task the DatasetAgent with two types of objectives: (1) expanding existing datasets, and (2) building entirely new datasets from scratch. For each task, we provide only high-level instructions, including the task type, the required number of categories, the total number of target images, and some general quality and semantic constraints. Notably, we do not supply any predefined image sources, allowing the DatasetAgent to autonomously collect, filter, and construct the datasets.

\noindent \textbf{The Dataset Expansion Task.} We use CIFAR-10 \cite{krizhevsky2009learning} and STL-10 \cite{coates2011analysis} as the base datasets for image classification tasks. The DatasetAgent is responsible for expanding each dataset by two orders of magnitude. The detailed statistics are shown in Table \ref{tab:dataset}, where $(1)$ denotes the first level of expansion and $(2)$ represents a larger-scale expansion. For object detection tasks, we select PASCAL VOC 2007 \cite{pascal-voc-2007} as our foundation dataset. Similarly, for image segmentation tasks, we utilize CamVid \cite{brostow2008segmentation} as our baseline dataset. 

\noindent \textbf{The Dataset Creation-From-Scratch Task.} The agent constructs two new datasets according to a specified 10-class schema, named Dataset-Ours(1) and Dataset-Ours(2), with the corresponding construction details also shown in Table \ref{tab:dataset}, and evaluates their performance on image classification tasks. For object detection and image segmentation tasks, we referred to MS-COCO to create a small general dataset only for verifying the effectiveness of DatasetAgent and tried to maintain the balance of categories as much as possible, and the images utilized in our dataset are curated from multiple existing large-scale datasets, such as the Mapillary Vistas Dataset \cite{neuhold2017mapillary} and are also derived from video clips and online sources. The dataset has four kinds of annotations, namely bounding box annotation for object recognition, pixel-level annotation for instance segmentation, category label annotation for semantic segmentation, and comprehensive annotation for panoramic segmentation. To better distinguish them, we named the datasets Obj-Dataset-Ours and Seg-Dataset-Ours respectively, in order to show their performance on different tasks. This targeted approach ensures our synthetic datasets address real-world challenges while maintaining practical relevance for domain-specific applications. 

\noindent \textbf{The Downstream Evaluation Stage.} The constructed datasets are used to train a set of widely adopted computer vision models to assess their quality and generalization ability across different tasks. For image classification, we employ: VGG-11 \cite{simonyan2015deepconvolutionalnetworkslargescale}, VGG-16 \cite{simonyan2015deepconvolutionalnetworkslargescale}, ResNet-18 \cite{he2015deepresiduallearningimage}, DenseNet-121 \cite{huang2018denselyconnectedconvolutionalnetworks}, GoogLeNet \cite{szegedy2014goingdeeperconvolutions}, EfficientNet-b0 \cite{tan2020efficientnetrethinkingmodelscaling}, ViT-B/16 \cite{dosovitskiy2021imageworth16x16words} and ViT-S/16 \cite{dosovitskiy2021imageworth16x16words}. For object detection evaluation, we utilize YOLO v3 \cite{redmon2018yolov3}, YOLO v8 \cite{varghese2024yolov8}, RetinaNet \cite{lin2017focal}, Fast R-CNN \cite{girshick2015fast}, and R-CNN \cite{girshick2014rich}. Our image segmentation evaluation comprises several specialized architectures: semantic segmentation models (FCN \cite{long2015fully}, U-Net \cite{ronneberger2015u}, DeepLab V2 \cite{chen2017deeplab}), instance segmentation models (Mask R-CNN \cite{he2017mask}, BlendMask \cite{chen2020blendmask}, CenterMask \cite{lee2020centermask}), and panoptic segmentation models (AUNet \cite{sun2020aunet}, MaskFormer \cite{cheng2021per}, Panoptic SegFormer \cite{li2022panoptic}). These models are independently trained on both the original and expanded datasets, and the test results are compared to validate the effectiveness of DatasetAgent. The newly built datasets Dataset-Ours(1), Dataset-Ours(2), Obj-Dataset-Ours and Seg-Dataset-Ours are also evaluated using the same model suite to analyze their performance. For fair comparison in downstream experiments, we utilized PyTorch's official pre-trained models across all comparative methods, maintaining identical hyperparameters and training cycles while employing the Adam optimizer throughout all experiments.

\begin{table*}[h]

  \caption{Image classification datasets and experimental results from DatasetAgent Construction. In this experiment, DatasetAgent utilizes LLaVA as the MLLM, and DeepSeek-R1 as the reasoning LLM.}
  \label{tab:dataset}
  \begin{tabular}{cccccccccccc}
    \toprule    
    \textbf{Dataset} &\textbf{Label} &\textbf{Original} &\textbf{Collected} &\textbf{New} &CBI&SSIM&ALR&DSE&SDI&DDC\\
    \hline
    \multirow{2}*{CIFAR-10} 
    &\multirow{2}{4.5cm}{Airplane, Automobile, Bird, Cat, Deer, Dog, Frog, Horse, Ship, Truck} 
    & \multirow{2}*{60000} & $(1)$ 12859 & 70188 &0.021 &0.943&99.8\%&2.78&0.267& 0.031\\
    &&&(2) 18259 & 75101&0.017&0.955&98.6\%&2.96&0.285&0.024\\
    \midrule
    \multirow{2}*{STL-10}&\multirow{2}{4.5cm}{Airplane, Car, Bird, Cat, Deer, Dog, Monkey, Horse, Ship, Truck}
    & \multirow{2}*{13000}& $(1)$ 13667 &25982 &0.009&0.963&99.2\%&2.85&0.278&0.018\\
    & & & (2) 19667&29832 &0.013&0.967&98.9\%&2.91&0.291&0.027\\
    \toprule
    \multirow{2}*{Dataset-Ours$(1)$} 
    &\multirow{2}{4.5cm}{Computer, Elephant, Panda, Penguin, Phone, Iris, Squirrel, Tiger, Train, Zebra} 
    & \multirow{2}*{0} & \multirow{2}*{13500} & \multirow{2}*{12964} 
    & \multirow{2}*{0.010}& \multirow{2}*{0.977}& \multirow{2}*{98.8\%}& \multirow{2}*{2.90}& \multirow{2}*{0.277}& \multirow{2}*{0.019}\\
    & & & & &&&&&&\\
    \hline
    \multirow{2}*{Dataset-Ours$(2)$} 
    &\multirow{2}{4.5cm}{Banana, Camel, Koala, Lemon, Llama, Mushroom, Orange, Peacock, Ball, Car} 
    & \multirow{2}*{0} & \multirow{2}*{13500} & \multirow{2}*{12756} 
    & \multirow{2}*{0.018}& \multirow{2}*{0.968}& \multirow{2}*{99.6\%}& \multirow{2}*{2.81}& \multirow{2}*{0.283}& \multirow{2}*{0.021}\\
    & & & & &&&&&&\\
 \toprule
\end{tabular}
\end{table*}

\begin{table*}[h]
  \caption{Results of downstream image classification models trained by the constructed datasets. }
  \label{tab:deep leaning}
  \centering
  \begin{tabular}{c|cccccccc|cc}
    \toprule
       \multicolumn{1}{c}{} 
       &\multicolumn{8}{|c|}{\textbf{Expansion-Accuracy (\%)}}  
       & \multicolumn{2}{c}{\textbf{Creation-Accuracy (\%)}}\\
    \cline{2-11}
   \multicolumn{1}{c}{} 
   & \multicolumn{2}{|c}{CIFAR-10 (1)} 
   & \multicolumn{2}{c}{CIFAR-10 (2)} 
   & \multicolumn{2}{c}{STL-10 (1)} 
   & \multicolumn{2}{c|}{STL-10 (2)} 
   & \multicolumn{1}{c}{Dataset-Ours} 
   & \multicolumn{1}{c}{Dataset-Ours}\\
     \cline{2-11}
   \multicolumn{1}{c|}{} &  $Ori$ & $New$ & $Ori$ & $New$ & $Ori$ & $New$& $Ori$ & $New$ & $(1)$ & $(2)$\\
    \toprule
   \multirow{1}*{VGG-11}  &86.59&86.81 &86.59 &87.42& 89.31&89.47& 89.31&89.69 & 98.47 &95.07 \\
   
    \toprule
      \multirow{1}*{VGG-16} &87.18& 88.62&87.18 &88.77& 90.97&91.05& 90.97&91.29 & 98.62 &95.70 \\
      
  \toprule
      \multirow{1}*{ResNet-18} &90.65& 91.64&90.65 &91.77& 89.30&90.64& 89.30&91.29 & 98.77 &97.00\\
     
   \toprule
  \multirow{1}*{Densenet121} &85.81& 85.89&85.81 &86.55& 88.51&89.01& 88.51&89.49 & 98.69 &96.84\\
  
      \toprule
  \multirow{1}*{ViT-S/16} &95.69&96.35&95.69 &96.61& 97.17&97.51& 97.17&\textbf{98.07} & 99.63 &98.44\\
 
  \bottomrule
  \multirow{1}*{ViT-B/16} &97.07&\textbf{ 97.11}&97.07 &\textbf{97.28}& 97.82&\textbf{97.69}& 97.82&98.01 & \textbf{99.78} &\textbf{98.67}\\
 
  \bottomrule
  \multirow{1}*{GoogLeNet} &89.69& 87.59&89.69 &88.31& 85.35&85.66& 85.35&86.01 & 98.55 &95.01\\
 
  \bottomrule
  \multirow{1}*{EfficientNet-b0} &89.69&89.75&89.69&89.73& 89.60&90.95& 89.60&92.10 & 98.70 &97.44\\
 
  \bottomrule
\end{tabular}
\end{table*}

All experiments are conducted on an NVIDIA GeForce RTX 4090 GPU. For multimodal reasoning, we employ LLaVA; for unimodal vision tasks, DeepSeek-R1 is used. For object detection we chose Grounding Dino 1.5 and Grounded-SAM \cite{ren2024grounded} for annotation and mask generation in the segmentation task. In ablation studies, alternative foundation models such as Qwen are also employed, all accessed via API interfaces. For consistency and fair comparison, identical training strategies were maintained across all experiments when training the same model on different datasets. In Table \cref{tab:dataset,tab:dataset_VOC,tab:dataset_CamVid}, the Label column denotes the category name; Original indicates the number of images in the original dataset for that class (set to 0 for newly created datasets); Collected refers to the total number of images retrieved; and New specifies the final number of images included in the constructed dataset after filtering by DatasetAgent.

\begin{table*}[h]
  \caption{Object Detection Datasets and Experimental Results from DatasetAgent Construction. Compared to the configuration in Table \ref{tab:dataset}, this experiment additionally employed Grounding Dino 1.5 as the Vision-Language Model (VLM) and introduced three additional evaluation metrics specifically designed for object detection datasets.}
  \label{tab:dataset_VOC}
  \centering
  \small
  \renewcommand{\arraystretch}{2.6} 
  \begin{tabular}{l@{\hspace{6pt}}l@{\hspace{6pt}}l@{\hspace{3pt}}l@{\hspace{3pt}}l@{\hspace{3pt}}|@{\hspace{4pt}}c@{\hspace{4pt}}c@{\hspace{4pt}}c@{\hspace{4pt}}c@{\hspace{4pt}}c@{\hspace{4pt}}c@{\hspace{4pt}}c@{\hspace{4pt}}c@{\hspace{4pt}}c}
    \toprule    
    \textbf{Dataset} &\textbf{Classes} &\textbf{Original} &\textbf{Collected} &\textbf{Total} &CBI&SSIM&ALR&DSE&SDI&DDC&IDDE&BQI&OSR\\
    \hline
    \multirow{2}*{\shortstack{VOC2007\\(train \& val)}} 
    &\multirow{2}{3cm}{\footnotesize Person, Bird, Cat, Cow, Dog, Horse, Sheep, Plane, Bike, Boat, Bus, Car, Mbike, Train, Bottle, Chair, Table, Plant, Sofa, TV}
    & \multirow{2}*{5011} & (1)5231 & 10242 &0.023 &0.952&98.6\%&2.81&0.371&0.027&0.686&0.915&0.083\\
    & & & (2)6179 & 11190&0.021&0.924&98.1\%&2.93&0.411&0.021&0.562&0.906&0.128\\
    \hline
\multirow{2}*{Obj-Dataset-Ours} 
&\multirow{2}{3cm}{\footnotesize Person, Car, Aeroplane, Bus, Train, Truck, Boat, Traffic Light, Traffic Sign, Bird, Cat, Dog, Horse, Sheep, Cow, Elephant, Bear} 
& \multirow{2}*{0} & \multirow{2}*{2221} & \multirow{2}*{2221} 
&\multirow{2}*{0.013}&\multirow{2}*{0.928}&\multirow{2}*{97.8\%}&\multirow{2}*{2.85}&\multirow{2}*{0.405}&\multirow{2}*{0.022}&\multirow{2}*{0.550}&\multirow{2}*{0.895}&\multirow{2}*{0.135}\\
& & & & & & & & & & & & &\\
\bottomrule
\end{tabular}
\end{table*}

\begin{figure*}[h]
  \centering
  \includegraphics[width=\linewidth]{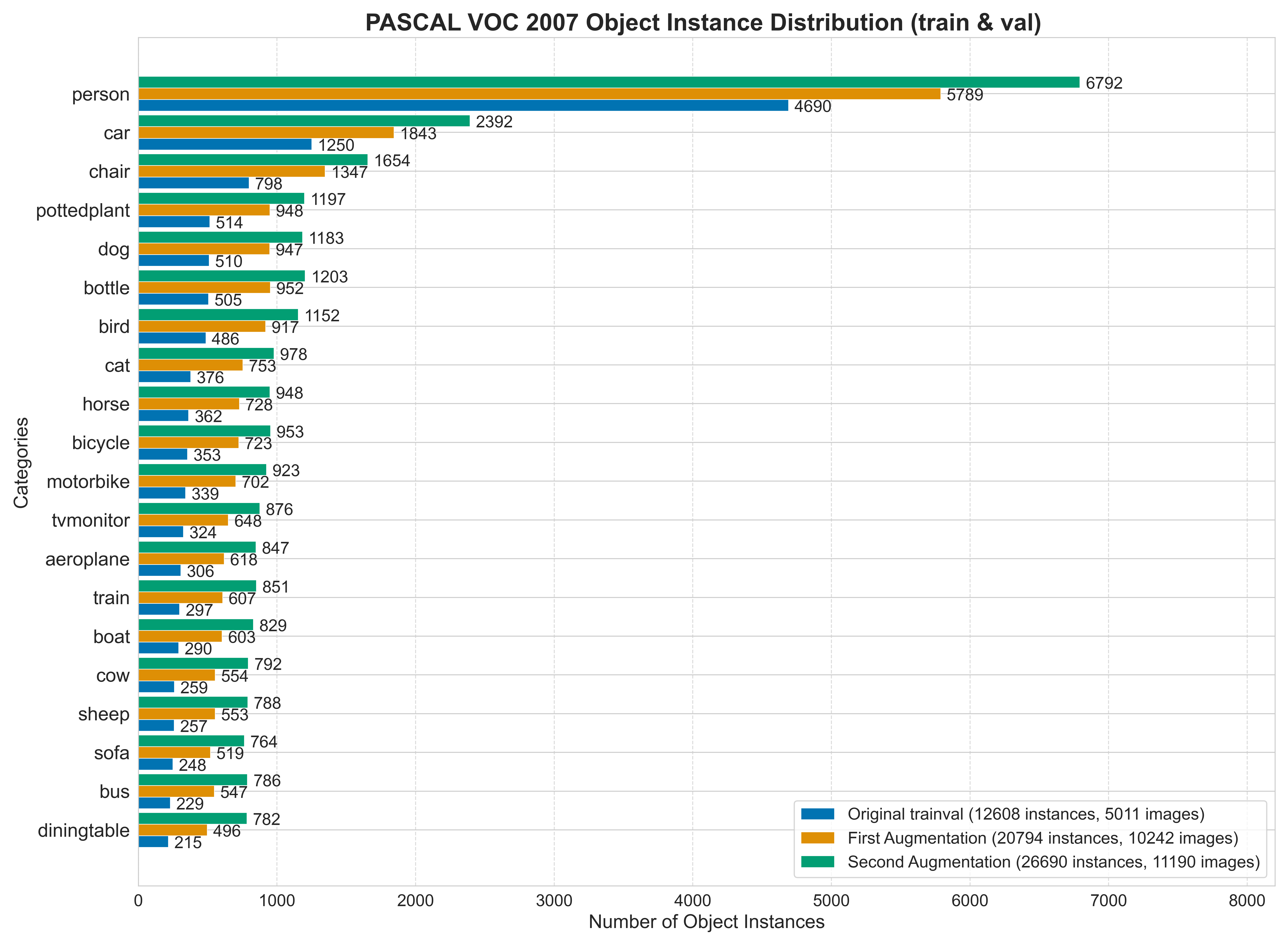} 
  \caption{A comparison of the number of instances and images of the two expansions on the PASCAL VOC2007 dataset. For this dataset, we choose to expand it for the training dataset and the validation dataset.}
  \label{fig:VOC2007}
\end{figure*}

\begin{table*}[h]
  \caption{Object Detection Performance on Original and Enhanced Datasets (mAP\%). Results are shown as mAP@0.5 / mAP@0.5:0.95. We selected several widely-used backbone architectures including ResNet \cite{he2016deep} and VGG \cite{simonyan2014very}.}
  \label{tab:detection_results}
  \centering
  \renewcommand{\arraystretch}{1.2}
  \begin{tabular}{lccccc}
    \toprule
    \textbf{Dataset} & \textbf{YOLOv3} & \textbf{YOLOv8} & \textbf{RetinaNet} & \textbf{Fast R-CNN} & \textbf{R-CNN} \\
    & \footnotesize{(416×416)} & \footnotesize{(640×640)} & \footnotesize{(ResNet-50)} & \footnotesize{(ResNet-101)} & \footnotesize{(VGG-16)} \\
    \midrule
    Original & 68.1 / 40.5 & 76.3 / 45.8 & 67.8 / 39.3 & 66.7 / 38.1 & 58.5 / 32.4 \\
    VOC2007(1) & 70.5 / 42.1 & 78.9 / 47.9 & 70.0 / 41.2 & 68.8 / 40.0 & 60.2 / 34.1 \\
    VOC2007(2) & 72.3 / 43.7 & 81.0 / 49.7 & 72.1 / 42.9 & 70.9 / 41.8 & 61.7 / 35.6 \\
    \midrule
    Obj-Dataset-Ours & 70.9 / 42.8 & 77.2 / 48.9 & 68.5 / 38.7 & 70.8 / 42.1 & 62.1 / 32.7 \\
    \bottomrule
  \end{tabular}
\end{table*}

\begin{figure*}[h]
  \centering
  \includegraphics[width=\linewidth]{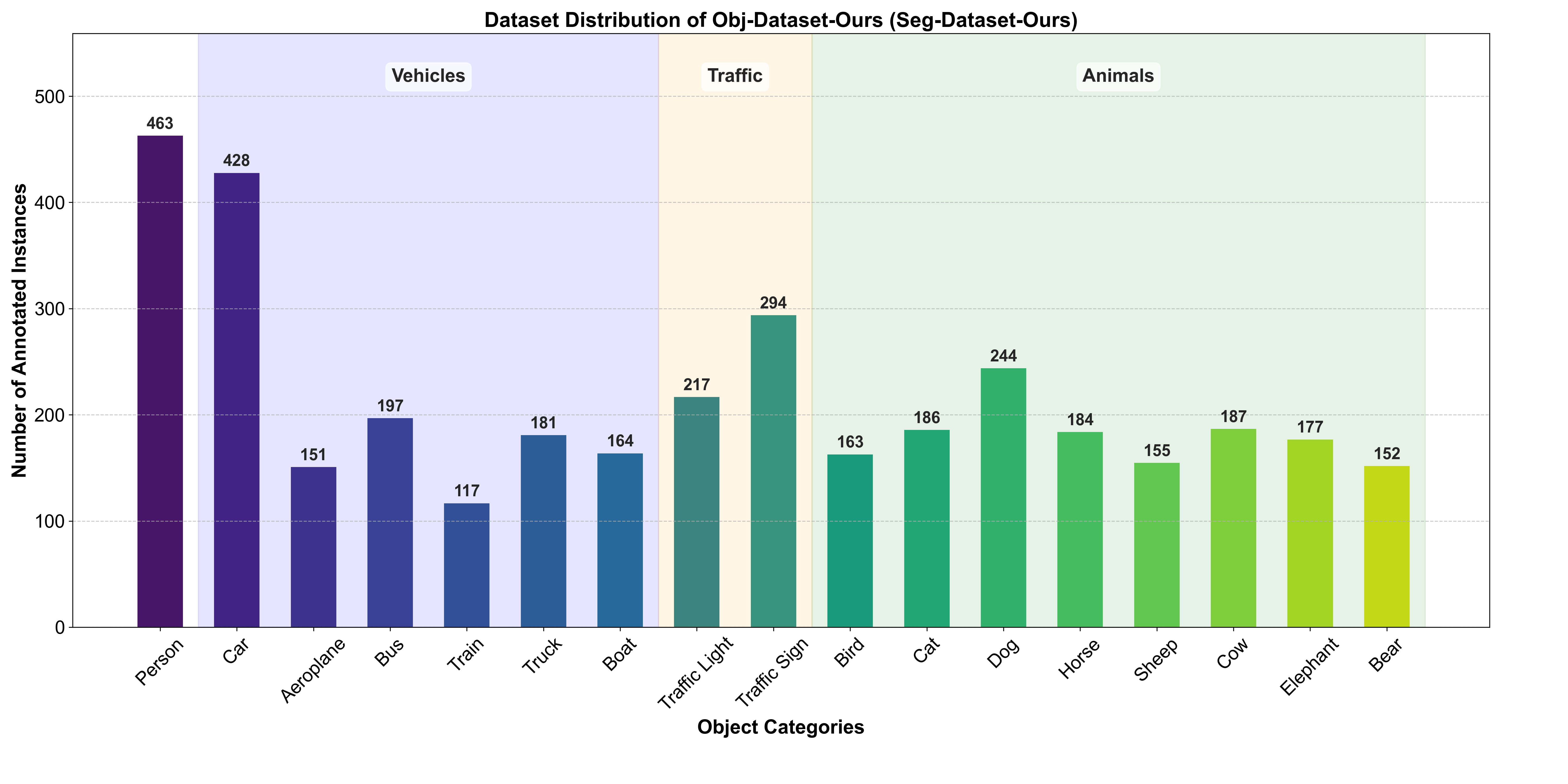}
  \caption{Dataset Distribution of Obj-Dataset-Ours (Seg-Dataset-Ours)}
  \label{fig:Ours}
\end{figure*}

\begin{table*}[h]
  \caption{DatasetAgent Construction Results on CamVid and Seg-Dataset-Ours. Note that Seg-Dataset-Ours and Obj-Dataset-Ours use the same images, so they are the same in the first six metrics. Similarly, in the construction of the dataset for image segmentation, DatasetAgent uses Grounded SAM for segmentation processing.}
  \label{tab:dataset_CamVid}
  \centering
  \small
  \renewcommand{\arraystretch}{2.5} 
  \begin{tabular}{l@{\hspace{6pt}}l@{\hspace{6pt}}l@{\hspace{3pt}}l@{\hspace{3pt}}l@{\hspace{3pt}}|@{\hspace{4pt}}c@{\hspace{4pt}}c@{\hspace{4pt}}c@{\hspace{4pt}}c@{\hspace{4pt}}c@{\hspace{4pt}}c@{\hspace{4pt}}c@{\hspace{4pt}}c@{\hspace{4pt}}c}
    \toprule    
    \textbf{Dataset} &\textbf{Classes} &\textbf{Original} &\textbf{Collected} &\textbf{Total} &CBI&SSIM&ALR&DSE&SDI&DDC&ESI&ACS&PCB\\
    \hline
    \multirow{2}*{CamVid} 
    &\multirow{2}{3.5cm}{\footnotesize Sky, Building, Pole, Road, Pavement, Tree, Symbol, Fence, Car, Pedestrian, Bicyclist, Void}
    & \multirow{2}*{701} & (1)536 & 1237 &0.028 &0.963&99.1\%&2.75&0.356&0.025&2.537&0.93&0.87\\
    & & & (2)552 & 1789 &0.020&0.935&98.5\%&2.89&0.392&0.023&2.871&0.91&0.83\\
    \hline
\multirow{2}*{Seg-Dataset-Ours} 
&\multirow{2}{3.5cm}{\footnotesize Person, Car, Aeroplane, Bus, Train, Truck, Boat, Traffic Light, Traffic Sign, Bird, Cat, Dog, Horse, Sheep, Cow, Elephant, Bear} 
& \multirow{2}*{0} & \multirow{2}*{2221} & \multirow{2}*{2221} 
&\multirow{2}*{0.025}&\multirow{2}*{0.918}&\multirow{2}*{97.8\%}&\multirow{2}*{2.85}&\multirow{2}*{0.405}&\multirow{2}*{0.022}&\multirow{2}*{2.58}&\multirow{2}*{0.89}&\multirow{2}*{0.85}\\
& & & & & & & & & & & & &\\
\bottomrule
\end{tabular}
\end{table*}

\subsection{Metric}
To comprehensively assess the quality and utility of the image classification datasets constructed by DatasetAgent, we adopt a set of representative quantitative evaluation metrics spanning key aspects such as class balance, image quality, annotation reliability, data source diversity, and feature diversity. \\
\textbf{Core Dataset Evaluation Metrics}
\begin{itemize}

\item \textbf{Class Balance Index (CBI)} evaluates label distribution uniformity by computing the standard deviation of class proportions: $\sigma = \sqrt{ \frac{1}{K} \sum_{i=1}^K (p_i - \frac{1}{K})^2 }$, where $p_i = \frac{n_i}{N}$ and $K$ is the number of classes. A lower $\sigma$ (e.g., < 0.1) indicates better balance. 

\item \textbf{Structural Similarity Index (SSIM)} quantifies structural fidelity via luminance, contrast, and structure terms: $SSIM(I, \hat{I}) = \frac{(2\mu_I \mu_{\hat{I}} + C_1)(2\sigma_{I\hat{I}} + C_2)}{(\mu_I^2 + \mu_{\hat{I}}^2 + C_1)(\sigma_I^2 + \sigma_{\hat{I}}^2 + C_2)}$. SSIM > 0.9 indicates high structural similarity. 

\item \textbf{Annotation Label Reliability (ALR)} is measured by manual inspection, with ALR > 95\% indicating accurate labeling and minimal semantic noise. 

\item \textbf{Data Source Entropy (DSE)} reflects source diversity, defined as $H = -\sum_{i=1}^N p_i \log_2 p_i$, where $p_i$ is the proportion of images from source $i$. Higher entropy implies more balanced sourcing. 

\item \textbf{Sample Diversity Index (SDI)} assesses intra-class feature diversity using average pairwise cosine similarity among feature vectors $\mathbf{f}i$:
$\text{SDI} = 1 - \frac{1}{n(n-1)} \sum{i=1}^{n} \sum_{j \neq i} s_{ij}$,
where $s_{ij} = \frac{\mathbf{f}_i \cdot \mathbf{f}_j}{|\mathbf{f}_i| |\mathbf{f}_j|}$. A higher SDI indicates greater semantic diversity and improved generalization potential. 

\item \textbf{Dataset Distribution Consistency (DDC)} uses the Kullback–Leibler divergence to quantify distribution shift in image classification dataset: $KL(P||Q) = \sum_i p_i \log(\frac{p_i}{q_i})$. A KL divergence below 0.1 suggests good consistency.

\end{itemize}
Beyond these general metrics, we introduce specialized evaluation criteria for object detection and image segmentation tasks to address their unique characteristics and quality requirements.\\
\textbf{Object Detection Specific Metrics}
\begin{itemize}
\item \textbf{Instance Density Distribution Entropy (IDDE)} quantifies the diversity distribution of objects at different scales by calculating the entropy of small/medium/large object proportions:
$H_{scale} = -\sum_{s \in \{S,M,L\}} p_s \log p_s$,
where $p_s$ is based on COCO standard definitions (small: area < $32^2$, medium: $32^2$ to $96^2$, large: > $96^2$). Higher entropy values indicate more balanced scale distribution.

\item \textbf{Bounding Box Quality Index (BQI)} assesses annotation precision through manual inspection:
$BQI = \frac{\text{count}(IoU_{gt} > 0.7) + 0.5 \cdot \text{count}(0.5 < IoU_{gt} \leq 0.7)}{\text{total samples}}$.
A BQI value exceeding 0.9 indicates high-quality annotations, requiring strict guidelines for edge adherence and occlusion.

\item \textbf{Occlusion Scenario Coverage Rate (OSR)} measures the proportion of samples containing occlusion relationships:
$OSR = \frac{\sum_{i=1}^N \mathbb{I}(OccLevel_i > 0)}{N}$.
We establish quantitative occlusion levels (e.g., partial occlusion >30\%, severe occlusion >60\%) to reflect real-world detection challenges.
\end{itemize}

\textbf{Image Segmentation Specific Metrics}

\begin{itemize}
\item \textbf{Edge Sharpness Index (ESI)} evaluates the clarity of segmentation boundaries by calculating the average edge gradient magnitude:
$ESI = \frac{1}{|E|}\sum_{p \in E}||\nabla I(p)||_2$,
where $E$ represents the set of segmentation edge pixels. Higher ESI values reflect clearer boundary delineation.

\item \textbf{Annotation Consistency Score (ACS)} measures inter-annotator agreement using the Dice coefficient:
$ACS = \frac{2|A \cap B|}{|A| + |B|}$.
We  randomly select 100 samples for annotation by three domain experts, with an average ACS exceeding 0.85 indicating reliable segmentation.

\item \textbf{Pixel Category Balance (PCB)} extends class balance metrics to the pixel level:
$PCB = 1 - \sqrt{\frac{1}{K}\sum_{k=1}^K (r_k - \frac{1}{K})^2}$,
where $r_k$ represents the proportion of pixels belonging to class $k$. A PCB value above 0.8 suggests appropriate pixel-level class distribution.
\end{itemize}

\begin{figure*}[h]
  \centering
  \includegraphics[width=\linewidth]{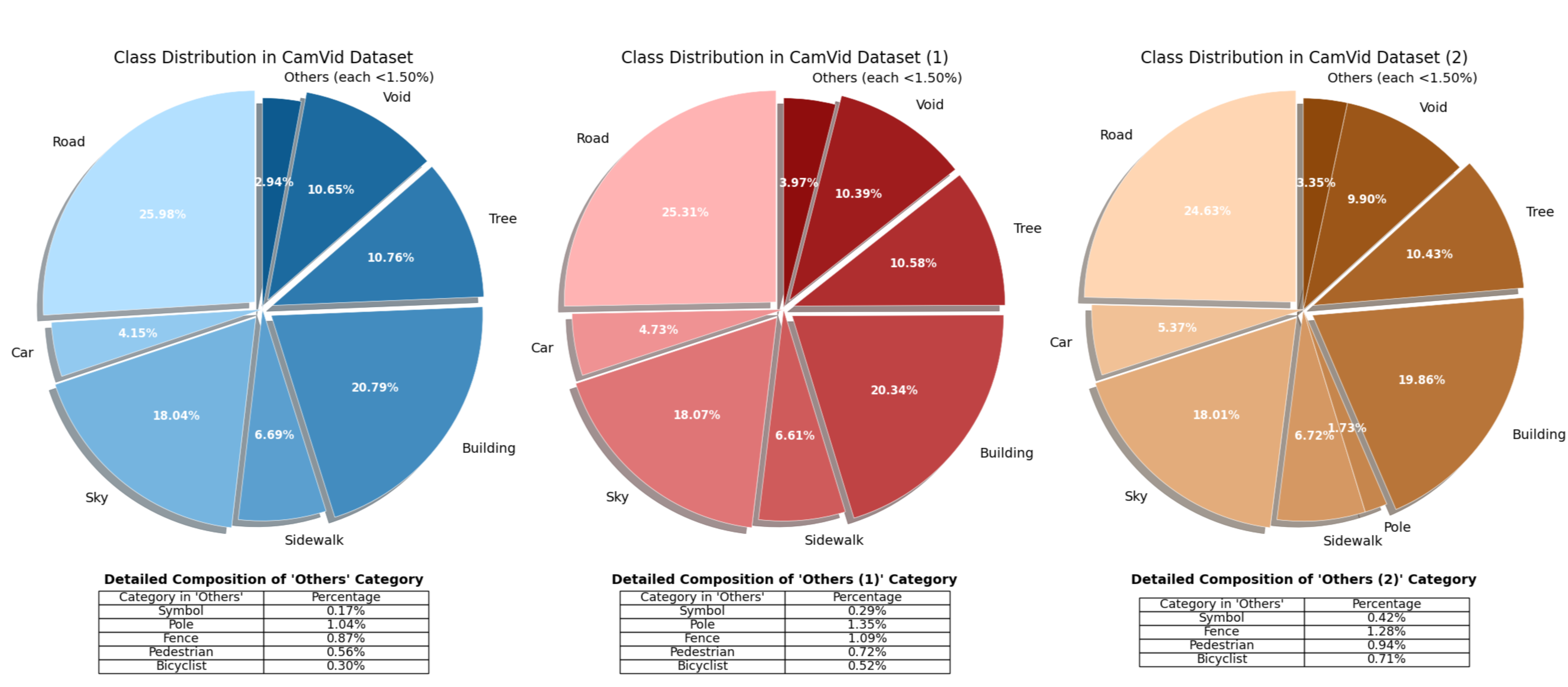}
  \caption{Instance distribution of CamVid with the two expansions. We adopted a coarse labeling approach, dividing CamVid instances into 11 specific categories and one empty category (containing all other objects), which facilitates more accurate statistics and annotations.}
  \label{fig:CamVid}
\end{figure*}

\subsection{Results for the Image Classification Task}

\textbf{Image Dataset Construction of Image Classfication.} As shown in Table \ref{tab:dataset}, the size of the original dataset, the number of collected images, and the size of the expanded dataset are displayed. To assess the overall performance of the DatasetAgent framework, we compare four datasets across six evaluation metrics, including both benchmark-augmented (CIFAR-10 and STL-10) and fully generated datasets (Ours(1), Ours(2)). The results demonstrate the high quality of image classification datasets constructed by DatasetAgent. Across all datasets, the Class Balance Index (CBI) remains low (0.009–0.021), with STL-10 variants exhibiting the most balanced label distributions, while the generated datasets also maintain near-uniform class proportions, indicating ability of DatasetAgent to enforce structural balance regardless of initialization. In terms of visual quality, all datasets achieve high SSIM scores (>0.94), with Dataset-Ours(1) reaching the peak (0.977), suggesting that even fully generated data can meet or exceed standard benchmarks in perceptual fidelity. Annotation quality, as measured by ALR, is consistently high (>98\%), with Dataset-Ours(2) achieving the best score (99.6\%), surpassing datasets derived from benchmark supervision. This confirms that the multi-stage LLM pipeline can produce semantically accurate labels without human annotation. In terms of diversity, both Data Source Entropy (DSE) and Sample Diversity Index (SDI) show competitive results across all datasets. STL-10(2) and Dataset-Ours(2) have the highest SDI (0.291 and 0.283, respectively), reflecting strong feature-level heterogeneity, which is critical for model generalization. Notably, the fully generated datasets achieve DSE values close to those of benchmark-based sets, indicating that DatasetAgent can effectively diversify image sourcing even without seed guidance. Finally, all datasets maintain low distributional KL divergence (DDC < 0.031), with Dataset-Ours(1) and CIFAR-10(2) showing the strongest internal consistency, highlighting the system’s ability to construct statistically coherent datasets at scale.
\\

\noindent \textbf{Performance of Image Classification Tasks.} Table \ref{tab:deep leaning} show the accuracy rates for eight deep learning models before and after expanding the CIFAR-10 and STL-10 datasets. For the expansion task, $(1)$ and $(2)$ represent two levels of expansion, with $Ori$ for the original dataset and $New$ for the expanded one, the bold number represents the maximum accuracy achieved from training the eight models on the same image dataset. In the dataset creation task, $(1)$ and $(2)$ refer to two curated ten-class datasets from ImageNet. Table \ref{tab:deep leaning} also shows the performance of these models on two created datasets. These models trained with the datasets constructed by DatasetAgent show an improved performance in image classification tasks. The average improvements are $0.52{\%}$ and $0.41{\%}$, with accuracy enhancements by two orders of magnitude. Besides these, Using DatasetAgent to create new datasets for training these eight models also yields good results. The average accuracy on the two datasets are $98.90{\%}$ and $96.77{\%}$, respectively.

\begin{table*}[h]
\centering
\caption{Performance of Segmentation Models on Original, Expanded CamVid Datasets and Seg-Dataset-Ours}
\label{tab:segmentation_models}
\begin{tabular}{l|c|ccc|c}
\hline
\multirow{2}{*}{Model} & \multirow{2}{*}{Backbone} & \multicolumn{3}{c|}{CamVid (\%)} & Our Dataset (\%) \\
\cline{3-6}
 &  & Ori & (1) & (2) & Seg-Dataset-Ours \\
\hline
\multicolumn{6}{c}{Semantic Segmentation Models - MIoU} \\
\hline
FCN & ResNet-50 & 65.9 & 67.2 & 68.9 & 65.7 \\
U-Net & Custom (512×512) & 66.8 & 69.3 & 71.1 & 67.4 \\
DeepLab V2 & ResNet-101 & 67.3 & 70.5 & 72.0 & 68.2 \\
\hline
\multicolumn{6}{c}{Instance Segmentation Models - mAP} \\
\hline
Mask R-CNN & ResNet-50-FPN & 36.2 & 37.4 & 39.6 & 39.3 \\
Mask R-CNN & ResNet-101-FPN & 37.5 & 40.8 & 41.8 & 41.8 \\
BlendMask & ResNet-101-FPN & 38.1 & 41.4 & 43.2 & 42.2 \\
CenterMask & ResNet-101-FPN & 38.4 & 40.7 & 42.8 & 41.5 \\
\hline
\multicolumn{6}{c}{Panoptic Segmentation Models - PQ} \\
\hline
AUNet & ResNet-50-FPN & 59.6 & 61.2 & 62.7 & 59.4 \\
MaskFormer & Swin-S & 60.5 & 61.4 & 63.2 & 61.1 \\
Panoptic SegFormer & ResNet-101 & 61.8 & 62.6 & 63.5 & 62.9 \\
\hline
\end{tabular}
\end{table*}

\subsection{Results for the Object Detection Task}
\textbf{Image Dataset Construction of Object Detection.}
Table \ref{tab:dataset_VOC} presents our dataset expansion results for PASCAL VOC2007 and the construction of Obj-Dataset-Ours using DatasetAgent. Our experiments confirm that DatasetAgent, supported by the Vision-Language Model, effectively builds high-quality object detection datasets as demonstrated by several key metrics. While addressing the inherent class imbalance challenges in object detection and image segmentation datasets, we achieved notable improvements across all evaluation dimensions. The Data Source Entropy (DSE) increased from 2.81 to 2.93 and the Sample Diversity Index (SDI) rose from 0.371 to 0.411, indicating enhanced information richness and conceptual diversity. Dataset cohesion similarly improved, with Dataset Distribution Coherence (DDC) decreasing from 0.027 to 0.021 and Instance Density Distribution Entropy (IDDE) reducing from 0.686 to 0.562. Most significantly, the Out-of-Sample Robustness (OSR) metric improved from 0.083 to 0.128, highlighting the expanded dataset's superior generalization capabilities for real-world applications. For Obj-Dataset-Ours, as shown in Figure \ref{fig:Ours}, the Class Balance Index (CBI) of 0.013 confirms our success in maintaining instance balance while preserving excellent performance across other evaluation metrics, further validating our dataset construction methodology.\\

\noindent \textbf{Performance on Object Detection Tasks.} Table \ref{tab:detection_results} presents mAP scores for five popular object detection models evaluated on original and enhanced datasets. For PASCAL VOC2007, we observe a consistent performance improvement across all models with progressive dataset enhancements. On YOLOv8 (640×640), the highest-performing model, accuracy improved from 76.3\% to 81.0\% (mAP@0.5) and from 45.8\% to 49.7\% (mAP@0.5:0.95) after the second expansion. Similar gains are evident across other architectures, with average improvements of 2.95\% for mAP@0.5 and 2.48\% for mAP@0.5:0.95 between the original dataset and PASCAL VOC2007(2). Remarkably, our Obj-Dataset-Ours, despite being created from scratch, demonstrates excellent performance due to its optimized volume and instance distribution characteristics. This is particularly evident in traditional architectures like Fast R-CNN, where it achieves 70.8\% / 42.1\% (mAP@0.5 / mAP@0.5:0.95). 

\subsection{Results for the Image Segmentation Task}
\textbf{Image Dataset Construction of Image Segmentation.} For the image segmentation task, we have not only expanded the quantity of CamVid images but also supplemented the annotation files and masks. This enables the dataset to support three types of segmentation tasks. Table \ref{tab:dataset_CamVid} presents a comparative analysis of semantic segmentation datasets with their corresponding quality metrics. The CamVid dataset, initially comprising 701 images across 12 semantic categories (including void), underwent two sequential expansion phases, resulting in 1237 and 1789 total images respectively. Each expansion phase demonstrates consistent improvements in semantic coherence as evidenced by the ALR metric and enhanced structural integrity reflected in the ESI values (2.537 to 2.871). Our independently constructed Seg-Dataset-Ours contains 2221 images spanning 17 diverse object categories, exhibiting quality metrics comparable to the expanded CamVid dataset, with notably strong performance in structural similarity (SSIM) for 0.918 and annotation logical rationality (ALR) for 97.8\%.\\

\noindent \textbf{Performance of Image Segmentation Tasks.} The experimental results, as summarized in the table, show the performance of segmentation models across tasks and datasets. For semantic segmentation, DeepLab V2 achieves the highest mean Intersection-over-Union (MIoU) of 72.0\% on the augmented CamVid (2), surpassing FCN (68.2\%) and U-Net (70.3\%). In instance segmentation, BlendMask demonstrates superior generalization with a maximum mean Average Precision (mAP) of 43.2\% on CamVid (2), outperforming Mask R-CNN variants. For panoptic segmentation, Panoptic SegFormer attains the highest Panoptic Quality (PQ) of 63.5\% on CamVid (2). Notably, expanded datasets consistently improve performance over their original versions, validating the effectiveness of Our work. Furthermore, models trained on our compact Seg-Dataset-Ours achieve competitive metrics across all tasks (e.g., 62.9\% PQ), demonstrating its practical efficacy despite limited scale. \\

\section{Conclusion and Future Work}

DatasetAgent is proposed as a multiagent system for auto-constructing image datasets. It achieves full automation and batch processing in construction of image datasets. 
It can automatically plan and execute all the steps for building a dataset, including image collecting, processing, and annotation based on the coordination of four different Agents. 
Furthermore, DatasetAgent 
addresses inefficiencies and the strain on human resources in manual dataset construction, while also overcoming limitations of using generated or synthetic images with real ones. 
In this way, DatasetAgent fills a critical gap in the application of AI Agents and LLMs for image dataset construction. 
Our experiments demonstrate the high quality of the constructed image datasets, which enhance downstream image classification model training, positioning DatasetAgent as a valuable tool for automated dataset construction. Currently, DatasetAgent can only construct image classification datasets for image classification tasks. 
In the future, we will explore more efficient methods to enhance the quality and speed of dataset construction, particularly for complex scene annotation. We'll also investigate DatasetAgent's potential in annotating domain-specific image datasets, such as medical images.

\appendix
\section*{Appendix}
\section{Large Language Model}
DatasetAgent utilizes two prominent large language models, namely Ollama's model \textbf{DeepSeek R1} and \textbf{Qwen 2.5}. Ollama represents a lightweight, task-oriented framework designed to enhance the capability of agents to decompose tasks and foster the development of multi-agent systems. It supports any Large Language Model (LLM) and has demonstrated significant efficacy in handling complex tasks.

\textbf{DeepSeek R1}, part of the Ollama ecosystem, represents an accessible and open Large Language Model (LLM) crafted for the purpose of enabling developers, researchers, and corporations to build, experiment with, and extend their generative AI concepts in a responsible manner. \textbf{Qwen 2.5}, another advanced model, exhibits cutting-edge competencies in the realms of general knowledge, adaptability, mathematical reasoning, tool application, and multilingual translation, positioning itself on par with the leading artificial intelligence models.
\href{https://ollama.com/library/deepseek-r1}{https://ollama.com/library/deepseek-r1}
\href{https://ollama.com/library/qwen2.5}{https://ollama.com/library/qwen2.5}

\textbf{LLaVA} is a novel end-to-end trained large multimodal model that combines a vision encoder and Vicuna for general-purpose visual and language understanding\cite{liu2023llava}. It is recognized for its impressive chat capabilities that mimic the spirit of the multimodal GPT-4 . Both of these models have the flexibility to be substituted with other large models as needed\cite{liu2023llava}. 
\href{https://ollama.com/library/llava}{https://ollama.com/library/llava}

Large models, constructed from deep neural networks with massive parameters and complex computational structures, possess billions or even trillions of parameters\cite{chang2024survey}. They are trained on vast amounts of data to learn intricate patterns and features, enabling the generation of human-like text and the answering of natural language questions\cite{minaee2024large}. This has enhanced the models' expressive and predictive capabilities, allowing them to handle more complex tasks and data with formidable power. 

It has multimodal understanding and multi-type content generation capabilities, perfectly combining big data, computational power, and algorithms, significantly enhancing the pre-training and generation capabilities of single-modality large models\cite{wu2023next}, as well as their application in multimodal and multi-scenario contexts. Since then, numerous Multimodal Large Language Models (MLLMs) have emerged\cite{li2024multimodal}, such as Llava, NeXT-GPT, LEGO, etc\cite{wu2023next,liu2023llava}. These MLLMs can process various types of data, such as text, images, audio, and other multimodal data\cite{li2024multimodal}. They integrate the capabilities of NLP and CV to achieve comprehensive understanding and analysis of multimodal information, enabling a more holistic comprehension and processing of complex data. Nowadays, MLLMs are applied in professional fields like autonomous driving and medical assistance diagnostics\cite{king2023future,digiorgio2023artificial}.

With the development of multimodal large models, an intelligent entity known as AI Agent has been born\cite{zeng2023agenttuning}. These agents, driven by large models, allow people to interact through natural language, executing and handling professional or complex work tasks with high automation\cite{wang2024survey,xi2023rise}. They possess autonomy, capable of making independent decisions based on the information perceived from the environment, such as vision and hearing, and carrying out corresponding physical or virtual actions to achieve human-set goals, replacing humans in certain tasks and greatly releasing human effort to improve work efficiency in various fields\cite{wang2024survey,xi2023rise,talebirad2023multi,wu2023autogen,liu2024agentlite,liu2023agentbench,li2024personal}.

\section{The performance of MLLMs in DatasetAgent}

On the four datasets of benchmarks, we conducted experiments using different MLLMs to compare the scores of DatasetAgents using different MLLMs. The CIFAR-10 and STL-10 scores are the average of the scores obtained by expanding the data set to two magnifications. The experimental results show that the scores of DatasetAgents using different MLLMs differ by no more than 2.0, fully reflecting the model portability of DatasetAgents, and the results are not affected by different MLLMs.
As shown in Table \ref{tab:differentmodel}.
\begin{table*}[h]
    \centering

  \caption{Comparative analysis of the performance of DatasetAgent when utilizing different MLLMs.}
  \label{tab:differentmodel}
  \begin{tabular}{c|cccc}
    \toprule  
    \multirow{3}*{\textbf{Models}} &\multicolumn{4}{|c}{\textbf{Benchmarks}}  \\
    & CIFAR-10 &STL10 &ImageNet-10 (1)& ImageNet-10 (2)
    
       \\
       \cline{2-5}
    &Score(\%) &Score(\%) &Score(\%) &Score(\%)  \\
    
       \toprule
  \multirow{1}*{Llava} &93.7&92.8&94.1&92.5\\
    
    \multirow{1}*{DeepSeek-VL2} &95.1&94.2&93.8&94.2\\
    
    \multirow{1}*{Qwen2.5-VL} &94.7&92.9&93.7&92.8\\
    
    \multirow{1}*{Llama3.2} &94.3&93.4&93.2&92.4\\
    
    \toprule
       
\end{tabular}
\end{table*}

\section{Visualization Analysis}
\begin{figure}[h]
    \centering
    \includegraphics[width=1\linewidth]{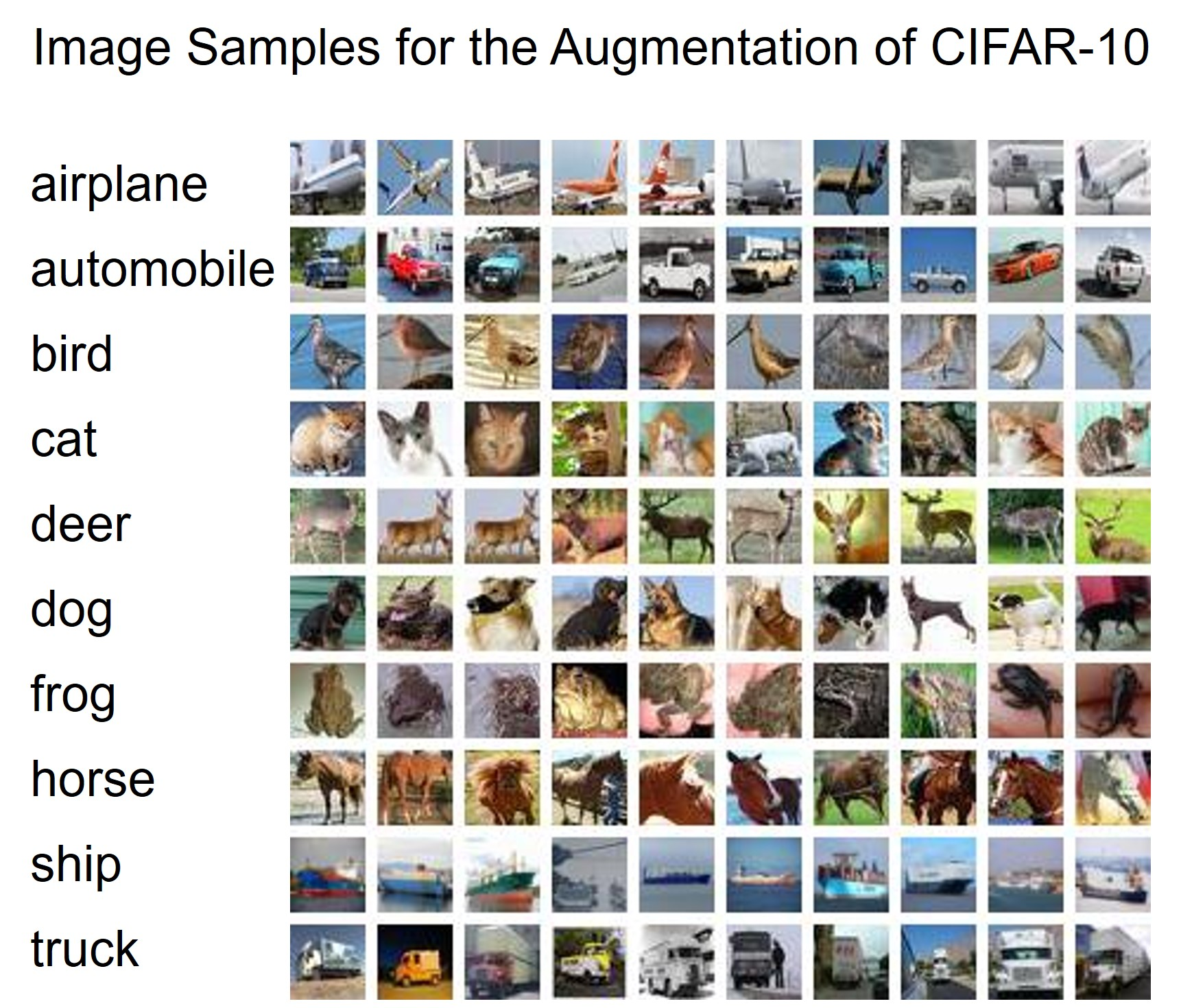}
    \caption{Real-world images processed by DatasetAgent for augmenting the CIFAR-10 dataset.}
    \label{fig:cifar10}
    \vspace{10pt} 
\end{figure}

Some visualization example data will be presented in this section.
\textbf{Image Classfication}
Taking the augmentation experiment of the CIFAR-10 (1) dataset as an example, the real-world images used for augmentation are illustrated in Figure \ref{fig:cifar10}. After being processed by the DatasetAgent, these images possess attributes that are fundamentally identical to those of the CIFAR-10 dataset.

\begin{figure}[h]
    \centering
    \includegraphics[width=1\linewidth]{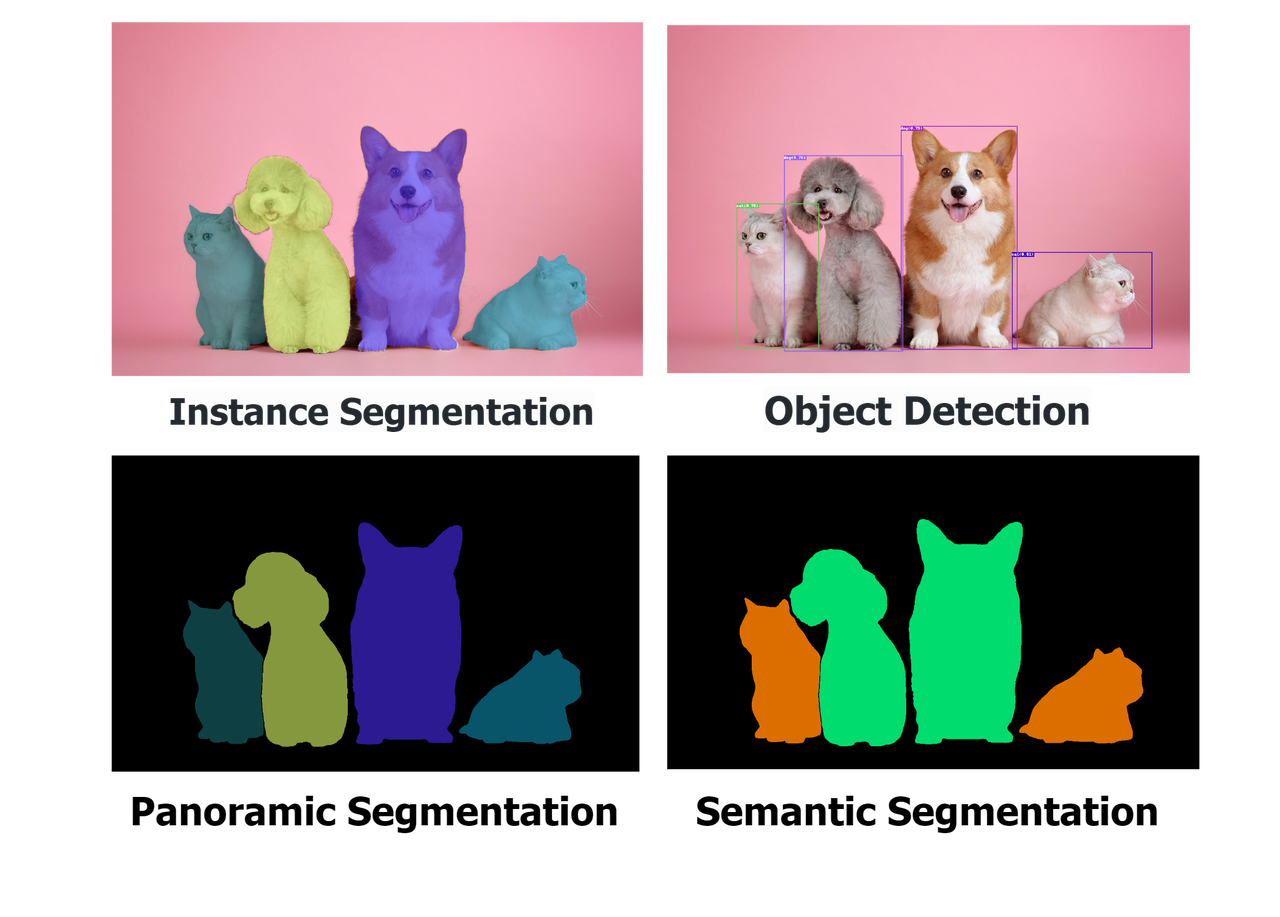}
    \caption{Four examples of image segmentation results.}
    \label{fig:visual}
    \vspace{10pt} 
\end{figure}

\textbf{Object Detection and Image Segmentation}
Figure \ref{fig:visual} shows the results of instance segmentation, object detection, panoramic segmentation, and semantic segmentation of the same example image after being processed by the DatasetAgent, where the VLM called is Ground SAM.

\section{Limitations}
DatasetAgent has demonstrated outstanding performance in image dataset construction, but we have found that there are still the following limitations:\\
1. When using large pre-trained models for interactive dialogue, analysis, annotation, and generation, it often requires higher operational costs and time, but this is still more efficient and cost-effective than traditional manual annotation methods.\\
2. The effectiveness of DatasetAgent is limited by the performance of large models, which is especially evident when using VLMs/LVMs for more complex image dataset task generation. We have found that when facing complex image environments or annotating uncommon objects, there are often still instances of poor annotation performance, resulting in an excessively high rejection rate of generated dataset samples. This is why manual annotation cannot be completely replaced at present.



\bibliography{main}

\end{document}